\documentclass[letterpaper, 10 pt, conference]{ieeeconf}  
\IEEEoverridecommandlockouts                              

\usepackage{amsmath}
\usepackage{graphicx}
\usepackage{amssymb}
\usepackage{threeparttable}
\usepackage{booktabs}
\usepackage{rotating}
\usepackage{balance}
\usepackage{xcolor}
\usepackage{url}
\usepackage[colorlinks=true, allcolors=red]{hyperref}
\usepackage[capitalise]{cleveref}

\title{\LARGE \bf
Universal Online Temporal Calibration for Optimization-based Visual-Inertial Navigation Systems
}

\author{Yunfei Fan$^{1}$, Tianyu Zhao$^{1}$, Linan Guo$^{2}$, Chen Chen$^{1}$, Xin Wang$^{1}$, Fengyi Zhou$^{1}$
\thanks{$1$ PICO Technology Co., Ltd., Beijing, China (e-mail: frank.01 $|$ zhaotianyu.1998 $|$ chenchen.pico $|$ billxin.wang $|$ finch.zhou@bytedance.com)}%
\thanks{$2$ China University of Mining and Technology (Beijing),
Beijing, China (e-mail: linanguo93@163.com)}%
}

\UseRawInputEncoding
\begin{document}

\maketitle
\thispagestyle{empty}
\pagestyle{empty}

\begin{abstract}
6-Degree of Freedom (6DoF) motion estimation with a combination of visual and inertial sensors is a growing area with numerous real-world applications. However, precise calibration of the time offset between these two sensor types is a prerequisite for accurate and robust tracking. To address this, we propose a universal online temporal calibration strategy for optimization-based visual-inertial navigation systems. Technically, we incorporate the time offset $t_d$ as a state parameter in the optimization residual model to align the IMU state to the corresponding image timestamp using $t_d$, angular velocity and translational velocity. This allows the temporal misalignment $t_d$ to be optimized alongside other tracking states during the process. As our method only modifies the structure of the residual model, it can be applied to various optimization-based frameworks with different tracking frontends. We evaluate our calibration method with both EuRoC~\cite{burri2016eurocdata} and simulation data and extensive experiments demonstrate that our approach provides more accurate time offset estimation and faster convergence, particularly in the presence of noisy sensor data.
\ The experimental code is available at \url{https://github.com/bytedance/Ts_Online_Optimization}.

\end{abstract}

\section{INTRODUCTION}
Estimating six-degree-of-freedom (6DoF) motion is fundamental to many real-world applications, including robotic systems, autonomous vehicles and tracking the ego-motion of virtual reality devices. In recent years, visual-inertial navigation systems (VINS), which utilize both visual signals from cameras and readings from inertial measurement units (IMUs), have become mainstream for their high accuracy and robustness. Compared with vision-based navigation systems, IMUs can help provide more accurate scale estimation and overcome problems with low texture or low illumination.

\begin{figure}[t]
        \centering
        \includegraphics[width=\linewidth]{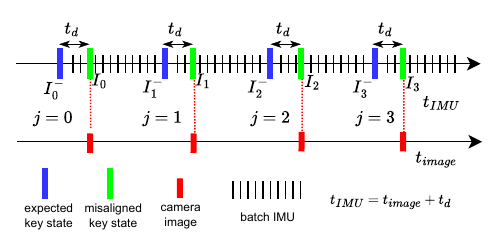}
        \caption{Timestamp misalignment in VINS. Using the $j^{th}$ image measurements to directly construct the residual model with the $j^{th}$ misaligned key state ($I_j$), ignoring the time offset $t_d$, will cause obvious inaccuracy because the $j^{th}$ image measurements are actually taken at the timestamp of the $j^{th}$ expected key state ($I^-_j$) considering the time offset $t_d$.} \label{fig:timestamp_misalignment}
\end{figure}

However, integrating two different types of sensors is challenging because they have different frequencies, time systems, and transmission delays. Incorrect temporal alignment can significantly degrade tracking accuracy and robustness or even lead to unacceptable divergence for VINS frameworks (see~\cref{fig:timestamp_misalignment}). Existing filter-based and optimization-based VINS frameworks model the time offset between the camera and IMU differently. For filter-based approaches, compelling MSCKF-based methods implicitly correct the time offset $t_d$ by integrating the cross-correlation between sliding window poses and $t_d$ with the covariance matrix. For optimization-based methods, Qin et al.~\cite{qin2018online} utilize optical flow to track the feature velocity and then compensate the feature observation under the camera timestamp to the IMU pose timestamp with the interframe feature velocity. However, this temporal calibration method is highly influenced by the precision of optical flow and is not feasible for VINS with different tracking frontends. On one hand, the calibration strategy is only activated when adequate landmarks are successfully tracked in consecutive frames. Adopting this strategy to low-cost image sensors with large noise may pose difficulties to the system robustness. On the other hand, VINS with other frontends like ORB-SLAM\cite{orbslam3} may not benefit from this method since the accurate computation of landmark velocity from pixel-level keypoint tracking is impossible.

To address the aforementioned challenges, we propose a universal online temporal calibration approach for optimization-based methods to perform time offset estimation and state estimation simultaneously. It does not rely on frontend tracking results so it can be easily applied to most VINS frameworks. Similar to Qin et al.~\cite{qin2018online}, our strategy models the time offset $t_d$ as a parameter in the system states to be continuously estimated. But different from them, we interpolate the IMU pose under the corresponding image's timestamp with the raw IMU pose, the latest $t_d$, as well as the estimated velocity and angular velocity to build the visual residual model (see ~\cref{fig:timestamp_misalignment}).
As a result, the Jacobian of the residual model naturally includes the sub-block with respect to $t_d$, allowing it to be properly optimized.
For the next sliding window optimization, the latest $t_d$ estimation is then used to align the timestamp range of observations. After several such iterations, the online estimation of $t_d$ is obtained.
Extensive experiments indicate that our proposed method can provide more accurate time offset estimations and 6-DoF ego motion estimations from highly noisy image data.

The contributions of this study are concluded as follows:
\begin{itemize}
        \item A universal online temporal calibration strategy for optimization-based VINS algorithms is proposed.
        \item The proposed approach is integrated into the existing VINS framework to validate its feasibility.
        \item Comprehensive experiments are conducted to assess its impact on both efficiency and accuracy. The experimental code is published to benefit community.
\end{itemize}

\section{RELATED WORK}
Over several decades, the problem of time offset calibration in VINS has garnered substantial attention. One prominent tool in this domain is the Kalibr calibration toolbox~\cite{furgale2013unified_kalibr}, which, along with methods such as that proposed by Kelly et al.~\cite{kelly2014general}, provides a solution for calibrating time offsets between different sensors. Nevertheless, these methods depend on a fixed planar calibration pattern, like a chessboard, and are confined to offline calibration scenarios.

To facilitate online calibration of time offsets in VINS, Li and Mourikis~\cite{li20133} introduced a seminal filter-based approach. This method integrates the cross-correlation between the poses within a sliding window and the time offset $t_d$ into the covariance matrix, thus enabling the implicit real-time correction of $t_d$ alongside pose estimation. Conversely, Qin et al.~\cite{qin2018online} proposed an optimization-based method explicitly addressing time axis misalignments. Their methodology adjusts observations by employing interframe feature velocity and time offset $t_d$, aligning these data points to their expected timestamps. However, a significant limitation of this approach is its reliance on accurate interframe feature velocity.
In contrast to the aforementioned methodologies, our proposed approach distinguishes itself through its universality. It avoids the need for specialized inputs like interframe feature velocity, thereby enhancing the adaptability across various VINS frameworks. This increased flexibility not only broadens the applicability of our method but also addresses limitations in prior approaches, offering a more robust solution for online time offset calibration.

\section{METHODOLOGY}
\subsection{IMU Factor}
In IMU-aided VINS system, each window would be mounted a key state $X_I$. In $j^{th}$ window, the key state and its error state are defined as Eq.~{(\ref{key_state})}, and Eq.~{(\ref{error_key_state})}, respectively.
\begin{align}
        \textbf{X}_{I_j} &= 
        \begin{bmatrix}
                \textbf{q}_{I_j}^\mathsf{T} ,{^G\textbf{p}_{I_j}^\mathsf{T}} ,{^G\textbf{v}_{I_j}^\mathsf{T} },\textbf{b}_{a_j}^\mathsf{T} ,\textbf{b}_{g_j}^\mathsf{T}      
        \end{bmatrix}^\mathsf{T} \label{key_state}
        \\
        \tilde{\textbf{X}}_{I_j} &= 
        \begin{bmatrix}
                \tilde{\boldsymbol{\theta}}_{I_j}^\mathsf{T}, 
                {^G\tilde{\textbf{p}}_{I_j}^\mathsf{T}}, 
                {^G\tilde{\textbf{v}}_{I_j}^\mathsf{T}}, 
                \tilde{\textbf{b}}_{a_j}^\mathsf{T}, 
                \tilde{\textbf{b}}_{g_j}^\mathsf{T}      
        \end{bmatrix}^\mathsf{T}  \label{error_key_state}
\end{align}
where $\textbf{q}_{I_j}$ denotes the rotation quaternion from the $j^{th}$ IMU frame $\{I\}$ to the world frame $\{G\}$. ${^G\textbf{p}_{I_j}}$ and ${^G\textbf{v}_{I_j}}$ denote the position and velocity of the $j^{th}$ IMU, respectively. $\textbf{b}_{a_j}$ and $\textbf{b}_{g_j}$ are the biases of the accelerometer and gyroscope, respectively.
Similar to~\cite{qin2018vinsmono}, if a batch of IMU measurements between the $j^{th}$ and $(j+1)^{th}$ frame (or keyframe) is successfully collected, the pre-integration is conducted on the batch IMU measurements using Eq.~(\ref{eqn-preintegration}).
\begin{equation}
        \begin{aligned}
                \boldsymbol{\alpha}^{I_j}_{I_{j+1}} &=\iint_{t\in[t_j,t_{j+1}]}{\textbf{R}^{I_j}_t(\hat{\textbf{a}}_t - \textbf{b}_{a_t}- \textbf{n}_a)}dt^2        \\
                \boldsymbol{\beta}^{I_j}_{I_{j+1}}  &=\iint_{t\in[t_j,t_{j+1}]}{\textbf{R}^{I_j}_t(\hat{\textbf{a}}_t - \textbf{b}_{a_t}- \textbf{n}_a)}dt        \\
                \boldsymbol{\gamma}^{I_j}_{I_{j+1}}  &=\int_{t\in[t_j,t_{j+1}]}{\frac{1}{2} \boldsymbol{\Omega}(\hat{\boldsymbol{\omega}}_t - \textbf{b}_{g_t}- \textbf{n}_{g})\gamma^{I_j}_t}dt        \\
        \end{aligned} \label{eqn-preintegration}
\end{equation}
where $\boldsymbol{\alpha}^{I_j}_{I_{j+1}}$, $\boldsymbol{\beta}^{I_j}_{I_{j+1}}$, $\boldsymbol{\gamma}^{I_j}_{I_{j+1}}$ are the position, velocity and attitude pre-integrations with respect to the period between $t_j$ and $t_{j+1}$. $\hat{a}_t$ and $\hat{\omega}_t$ are the raw IMU measurements (comprised of accelerometer and gyroscope data). $b_{a_t}$ and $b_{g_t}$ are the biases of the gyroscope and accelerometer, respectively, and $n_a$ and $n_g$ are their respective measurement noises. $\Omega{(\_)}$ can yield the corresponding quaternion. The pre-integration above represents the inertial constraint between the $j^{th}$ and $(j+1)^{th}$ frames (or keyframes). The corresponding inertial residual is as given in Eq.~{(\ref{eqn-0})}.
\begin{equation}
\begin{aligned}
\textbf{r}_j &=
\begin{bmatrix}
        \boldsymbol{\alpha}^{I_j}_{I_{j+1}} \\
        \boldsymbol{\beta}^{I_j}_{I_{j+1}} \\
        \boldsymbol{\gamma}^{I_j}_{I_{j+1}} \\
        \textbf{0} \\
        \textbf{0} 
\end{bmatrix} - 
\begin{bmatrix}
        {\varGamma }_P \\
        \textbf{R}^{G\mathsf{T}}_{I_j}({^G\textbf{v}_{I_{j+1}}} + {^G{\mathbf{g}}} \Delta{t_j} - {^G\textbf{v}_{I_j}}) \\
        {{\textbf{q}^G_{I_j}}^{-1}} \otimes {{\textbf{q}^G_{I_{j+1}}}} \\
        \textbf{b}_{a_{j+1}} - \textbf{b}_{a_{j}} \\
        \textbf{b}_{g_{j+1}} - \textbf{b}_{g_{j}}
\end{bmatrix} \\ \label{eqn-0}
{\varGamma }_P &=\textbf{R}^{G\mathsf{T}}_{I_{j}}( {^G\textbf{p}_{I_{j+1}}} - {^G\textbf{p}_{I_j}} + \frac{1}{2} {^G{\mathbf{g}}} \Delta{t_j^2} - {^G\textbf{v}_{I_j}}\Delta{t_j})
\end{aligned}
\end{equation}
where $\textbf{R}^G_{I_j}$ is the rotation matrix from the $j^{th}$ $\{I\}$ to $\{G\}$ in the sliding window. $^G\textbf{p}_{I_j}$ and $^G\textbf{p}_{I_{j+1}}$ represent the position of the $j^{th}$ and $(j+1)^{th}$ $\{I\}$ in the sliding window, respectively. $^G\textbf{v}_{I_j}$ and $^G\textbf{v}_{I_{j+1}}$ represent the velocity of the $j^{th}$ and $(j+1)^{th}$ $\{I\}$ in the sliding window, respectively. ${^G{\mathbf{g}}}$ is the gravity of earth. $\Delta{t}_j$ is the time period from the $j^{th}$ $\{I\}$ to $(j+1)^{th}$ $\{I\}$ in the sliding window.
\subsection{General Visual Factor}
In traditional feature-based VINS algorithms, extracted features are tracked frame by frame. After feature tracking, coordinates of features estimated jointly before can be re-projected onto the camera plane to construct the re-projection error for the cost function. For more convenience or accuracy, there are two mainstream parameterization methods of feature coordinate: 3D position or depth/inverse depth relative to a specific anchor pose. In whats follow, visual factors using these two parameterization methods are introduced.

1) \textbf{3D Position Parameterization:} The feature coordinate is parameterized as 3D coordinate ($^G\textbf{P}_{f_k}=[{^GX_k}, {^GY_k}, {^GZ_k}]^\mathsf{T}$) in the global frame. The visual factor is commonly defined as the re-projection error as given in Eq.~{(\ref{eqn-1})}
\begin{equation}
\begin{aligned}
        \textbf{r}_{j,k} &= \textbf{z}_{j,k} - \pi{({\textbf{R}^G_{I_j}}{\textbf{R}^I_C}(^G\textbf{P}_{f_k}-{^G{\textbf{P}_{I_j}}}-{\textbf{R}^G_{I_j}}{^{I_j}{\textbf{P}_C}}))} \\
        \textbf{z}_{j,k} &= [u_{j,k},v_{j,k}]^\mathsf{T}
\end{aligned} \label{eqn-1}
\end{equation}
where $\textbf{r}_{j,k}$ is the visual residual of the $k^{th}$ feature with respect to the $j^{th}$ $\{C\}$ in the sliding window. $\textbf{z}_{j,k}$ is the undistorted observation of $k^{th}$ feature measured in $j^{th}$ frame. ${\textbf{R}^I_C}$ and ${^I{\textbf{P}_C}}$ are the extrinsic parameters denoting rotation and translation from camera frame $\{C\}$ to IMU frame $\{I\}$, respectively. ${\textbf{R}^G_{I_j}}$ and ${^G\textbf{P}_{I_j}}$ are the rotation and  translation from IMU frame $\{I\}$ of $j^{th}$ state of the sliding window to global frame $\{G\}$. $\pi{(\_)}$ denotes the camera re-projection model, by which the 3D feature is re-projected into the image plane using the camera intrinsic parameters.

2) \textbf{Depth Parameterization:} The feature coordinate is parameterized as 脱水either depth or inverse depth with respect to the anchor camera frame. For example, the $\lambda_{k,i}$ represents the inverse depth of the $k^{th}$ feature on the $i^{th}$ anchor camera frame. The visual factor is defined as the residual with Eq.~{(\ref{eqn-2})}.
\begin{equation}
        \begin{aligned}
                \textbf{r}_{j,k} &= \textbf{z}_{j,k} - \pi{({\textbf{R}^G_{I_j}}{\textbf{R}^I_C}(({^G\textbf{P}_{f_k}}-{^G{\textbf{P}_{I_j}}}-{\textbf{R}^G_{I_j}}{^{I_j}{\textbf{P}_C}}))} \\
                \textbf{z}_{j,k} &= [u_{j,k},v_{j,k}]^\mathsf{T}
        \end{aligned} \label{eqn-2}
\end{equation}
where, the 3D feature coordinate $^{G}\textbf{P}_{f_k}$ is formulated with the inverse depth $\lambda_{k,i}$ and its anchor camera pose  
$\begin{bmatrix}
        {\textbf{R}^G_{I_i}} & {^G{\textbf{P}_{I_i}}}
\end{bmatrix}$.
\begin{equation}
        \begin{aligned}
                {^{G}\textbf{P}_{f_k}} &= {\textbf{R}^G_{I_i}}{\textbf{R}^I_C} {^{C_i}\textbf{P}_{f_k}} + {^G{\textbf{P}_{I_i}}}+{\textbf{R}^G_{I_i}}{^{I}{\textbf{P}_C}}    \\
                        {^{C_i}\textbf{P}_{f_k}} &= {\lambda_{k,i}^{-1}}
                        \begin{bmatrix} 
                                \textbf{z}_{i,k} \\ 
                                1 
                        \end{bmatrix}
        \end{aligned}\label{eqn-2.1}
\end{equation}
where $\textbf{z}_{i,k}$ denotes the undistorted measurement of $k^{th}$ feature on $i^{th}$ camera. 

\subsection{Model Construction for Temporal Calibration}

\begin{figure}[t]
        \centering
        \includegraphics[width=\linewidth]{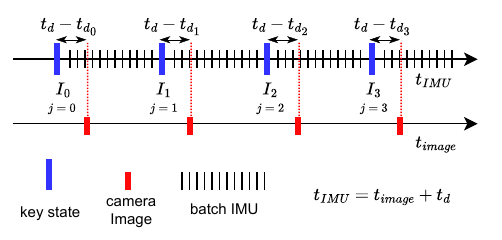}
        \caption{Timestamp misalignment in VINS. In the sliding window, the reference time axis is defined with IMU time axis. When the $j^{th}$ image is coming, the key state in sliding window at ($t_{image} + t_{d_{j-1}}$) will be built and this key state will be mounted with this camera measurements to reduce the impact of timestamp misalignment as possible in raw data assignment.} 
        \label{fig:timestamp_misalignment_compensation}
\end{figure}
To address timestamp misalignment, the timestamp compensation parameter $t_d$ is formulated into the system state for joint estimation with other states. With the traditional strategy of maintaining feature coordinates in the system state, the formulation of the system state $X$ is defined as Eq.~{(\ref{eqn-8})} for 3D position parameterization and Eq.~{(\ref{eqn-8.1})} for inverse depth parameterization.
\begin{align}
        \textbf{X} &=
        \begin{bmatrix}
                {\textbf{X}_{I_0}^\mathsf{T}}, 
                {\textbf{X}_{I_1}^\mathsf{T}}, 
                \dots,
                {\textbf{X}_{I_m}^\mathsf{T}}, 
                {^G\textbf{P}_{f_0}^\mathsf{T}}, 
                {^G\textbf{P}_{f_1}^\mathsf{T}},
                \dots,
                {^G\textbf{P}_{f_n}^\mathsf{T}}, {t_d}
        \end{bmatrix}^\mathsf{T}    \label{eqn-8}    
        \\ 
        {{\textbf{X}}} &=
        \begin{bmatrix}
                {{\textbf{X}}_{I_0}^\mathsf{T}}, 
                {{\textbf{X}}_{I_1}^\mathsf{T}}, 
                \dots,
                {{\textbf{X}}}_{I_m}^\mathsf{T}, 
                {{\lambda}}_{0,0}, 
                {{\lambda}}_{1,0}, 
                \dots,
                {{\lambda}_{n,0}} , {{{t_d}}}
        \end{bmatrix}^\mathsf{T} \label{eqn-8.1}    
\end{align}

To achieve precise state estimation including highly accurate temporal calibration, it is necessary to utilize the newest optimal $t_d$ to determine which image time interval is aligned with the incoming batch of IMU measurements (see~\cref{fig:timestamp_misalignment_compensation}). 
When constructing the visual factor using image measurements, two key steps must be handled. First, the camera pose at which the image related to the feature is captured must be provided. Second, there must be a connection established between the system state $t_d$ and the visual residual.
To handle the first step, the goal camera pose is compensated using the estimated $t_d$. A Jacobian containing a sub-block with respect to $t_d$ is then constructed between the residual and system state to build the connection of $t_d$ with respect to the residual. For the two mainstream parameterization methods, similar modifications are made to the residual formulations.

1) \textbf{3D Position Parameterization:} 
The primary modification is that the camera pose needs to be compensated for re-projection error.
\begin{equation}
\begin{aligned}
        \textbf{r}_{j,k} &= \textbf{z}_{j,k} - \pi{(\check{\textbf{R}}^G_{I_j}{\textbf{R}^I_C}({^G\textbf{P}_{f_k}}-{^G{\check{\textbf{P}}_{I_j}}}-{\textbf{R}^G_{I_j}}{^I{\textbf{P}_C}}))} \\
        \textbf{z}_{j,k} &= [u_{j,k},v_{j,k}]^\mathsf{T} \\
\end{aligned} \label{eqn-3}
\end{equation}
where 
$\begin{bmatrix}
        \check{\textbf{R}}^G_{I_j} & {^G{\check{\textbf{P}}_{I_j}}}
\end{bmatrix}$ is the compensated $j^{th}$ IMU pose (Rotation and translation) in sliding window which is used for re-projection. The compensation formulation is defined as Eq.~{(\ref{eqn-4})}
\begin{equation}
        \begin{aligned}
                \check{\textbf{R}}^G_{I_j} &={\textbf{R}^G_{I_j}}{\delta\textbf{R}_{I_j}}\\
                {^G{\check{\textbf{P}}_{I_j}}} &= {^G{\textbf{P}}_{I_j}} + {^G\textbf{v}_{I_j}}{\Delta{_{t_d}^j}} \\
                {\Delta{_{t_d}^j}} &= t_d - t_{d_j} \\
                \delta\textbf{R}_{I_j} &= {(\textbf{I}+[{\boldsymbol{\omega}_j}{\Delta{_{t_d}^j}}]_{\times})}
        \end{aligned} \label{eqn-4}
\end{equation}
where ${\boldsymbol{\omega}_j}$ is the angular velocity of the $j^{th}$ $\{I\}$. $t_{d_j}$ is the timestamp compensation used when constructing the $j^{th}$ window. Thus, as the $j^{th}$ window becomes older, $t_{d_j}$ becomes increasingly inaccurate. Furthermore, the corresponding Jacobian formulation is defined in Eq.~{(\ref{eqn-4.1})}
\begin{equation}
        \textbf{J}_{X,j,k} = 
        \textbf{J}_{j,k} 
        \begin{bmatrix}
                \textbf{J}_{I} & \textbf{J}_{f_k} & \textbf{J}_{t_d}
        \end{bmatrix} \label{eqn-4.1}
\end{equation}
where the sub-Jacobian is presented as Eq.~{(\ref{eqn-4.2})}
\begin{equation}
        \begin{aligned}
        \textbf{J}_{I}&=  
        \begin{bmatrix}
                \textbf{0}_{3\times15j} & \textbf{J}_{I_j} & \textbf{0}_{3\times15(m-j-1)}
        \end{bmatrix}
        \\
        \textbf{J}_{I_j}&= 
        \begin{bmatrix}
        {\textbf{R}^I_C}{^\mathsf{T}}
        \begin{bmatrix}
                ^{I_j}\check{\textbf{P}}_{f_k}
        \end{bmatrix}_{\times}{{\delta{\textbf{R}}}_{I_j}^\mathsf{T}}
                &
        -{\textbf{R}^I_C}{^\mathsf{T}}{{\check{\textbf{R}}}^G_{I_j}}{^\mathsf{T}}
        &
        \textbf{0}_{3\times9}
        \end{bmatrix} 
        \\
        \textbf{J}_{{f_k}}&= 
        \begin{bmatrix}
                \textbf{0}_{3\times{3k}} & {\textbf{R}^I_C}{^\mathsf{T}}{{\check{\textbf{R}}}^G_{I_j}}{^\mathsf{T}} & \textbf{0}_{3\times{3(n-k-1)}}
        \end{bmatrix}
        \\
        \textbf{J}_{t_d}  &= 
        {{\textbf{R}^I_C}{^\mathsf{T}}}
        \begin{bmatrix}
                ^{I_j}\check{\textbf{P}}_{f_k}
        \end{bmatrix}_{\times}{\boldsymbol{\omega}_j} - 
        {{\textbf{R}^I_C}{^\mathsf{T}}}
        {\check{\textbf{R}}^G_{I_j}}{^\mathsf{T}}{^G\textbf{v}_{I_j}} \\
        {^{I_j}\check{\textbf{P}}_{f_k}} &=      
        \check{\textbf{R}}^G_{I_j}{^\mathsf{T}}(
                {^G\textbf{P}_{f_k}}-
                {^G\check{\textbf{P}}_{I_j}})
        \end{aligned} \label{eqn-4.2}
        \end{equation}
Furthermore, for simplicity, the camera model is defined as a pinhole, and the resulting Jacobian $\textbf{J}_{j,k}$ of $\textbf{r}_{j,k}$ with respect to the position of the feature is defined in Eq.~{(\ref{eqn-4.3})}
\begin{equation}
        \begin{aligned}
                \textbf{J}_{j,k} &=
                \frac{1}{{^{C_j}{Z}_k}^2}
                \begin{bmatrix}
                {^{C_j}{Z}_k}&0&-{^{C_j}{X}_k}\\
                0&{^{C_j}{Z}_k}&-{^{C_j}{Y}_k}\\
             \end{bmatrix} \\
             {^{C_j}\textbf{P}_{f_k}} &=
             \begin{bmatrix}
                {^{C_j}X_{k}} &{^{C_j}Y_{k}} &{^{C_j}Z_{k}}
             \end{bmatrix}^\mathsf{T}
        \end{aligned}
     \label{eqn-4.3}
     \end{equation}
2) \textbf{Depth Parameterization:}
The primary modification is the compensation for both the camera pose utilized to conduct re-projection error and the anchor camera pose of the feature.
\begin{equation}
        \begin{aligned}
                \textbf{r}_{j,k} &= \textbf{z}_{j,k} - \pi{({\check{\textbf{R}}^G_{I_j}}{\textbf{R}^I_C}(({^G\check{\textbf{P}}_{f_k}}-{^G{\check{\textbf{P}}_{I_j}}}-{\check{\textbf{R}}^G_{I_j}}{^{I_j}{\textbf{P}_C}}))}
        \end{aligned} \label{eqn-5}
\end{equation}
where, the compensated pose ($\check{\textbf{R}}^G_{I_j}$, $^G\check{\textbf{P}}_{I_j}$) of the projection frame is calculated using Eq.~{(\ref{eqn-4})}. Since the 3D feature coordinate ${^G\check{\textbf{P}}_{f_k}}$ is affected by the $i^{th}$ anchor camera pose, it is obtained by using the compensated anchor camera pose and the depth $\lambda_{k,i}$. This is defined in Eq.~{(\ref{eqn-6})}
\begin{equation}
{^G\check{\textbf{P}}_{f_k}} = {\check{\textbf{R}}^G_{I_i}}{\textbf{R}^I_C}{^{C_i}\textbf{P}_{f_k}} + {^G{\check{\textbf{P}}_{I_i}}}+{\check{\textbf{R}}^G_{I_i}}{^{I}{\textbf{P}_C}} \label{eqn-6}
\end{equation}
where
\begin{equation}
        \begin{aligned}
        \check{\textbf{R}}^G_{I_i} &={\textbf{R}^G_{I_i}{\delta\textbf{R}_{I_i}}} \\
        {^G{\check{\textbf{P}}_{I_i}}} &= {^G{\textbf{P}}_{I_i}} + {^G\textbf{v}_{I_i}}{\Delta{_{t_d}^i}} \\
        {\Delta{_{t_d}^i}} &= t_d - t_{d_i} \\
        \delta\textbf{R}_{I_i} &= {(\textbf{I}+[{\boldsymbol{\omega}_i}{\Delta{_{t_d}^i}}]_{\times})}
        \end{aligned} \label{eqn-7}
        \end{equation}

Similar to 3D position parameterization, the corresponding Jacobian formulation of Eq.~{(\ref{eqn-6})} is also defined as Eq.~{(\ref{eqn-4.1})}. However, unlike Eq.~{(\ref{eqn-4.2})}, the definitions of these sub-Jacobians are shown in Eq.~{(\ref{eqn-7.2})}
\begin{equation}
\begin{aligned}
\textbf{J}_{I}&=  
\begin{bmatrix}
        \textbf{0}_{3\times15i} & \textbf{J}_{I_i} & \textbf{0}_{3\times15(j-i)} & \textbf{J}_{I_j} & \textbf{0}_{3\times15(m-j-1)}
\end{bmatrix}
\\
\textbf{J}_{I_i}&= 
\begin{bmatrix}
-{\textbf{R}^I_C}{^\mathsf{T}}{{\check{\textbf{R}}}^G_{I_j}}{^\mathsf{T}}{\textbf{R}^G_{I_i}}
\begin{bmatrix}
        {\delta\textbf{R}_{I_i}}{^{I_i}\check{\textbf{P}}_{f_k}}
\end{bmatrix}_{\times}
        &
{\textbf{R}^I_C}{^\mathsf{T}}{{\check{\textbf{R}}}^G_{I_j}}{^\mathsf{T}}
&
\textbf{0}_{3\times9}
\end{bmatrix} 
\\
\textbf{J}_{I_j}&= 
\begin{bmatrix}
{\textbf{R}^I_C}{^\mathsf{T}}
\begin{bmatrix}
        ^{I_j}\check{\textbf{P}}_{f_k}
\end{bmatrix}_{\times}
{\delta\textbf{R}_{I_j}^\mathsf{T}}
        &
-{\textbf{R}^I_C}{^\mathsf{T}}{{\check{\textbf{R}}}^G_{I_j}}{^\mathsf{T}}
&
\textbf{0}_{3\times9}
\end{bmatrix} 
\\
\textbf{J}_{f_k}&= 
\begin{bmatrix}
        \textbf{0}_{3\times{k}} & {-{{\lambda}_{k,i}^{-2}}\textbf{R}^I_C} & \textbf{0}_{3\times{(n-k-1)}}
\end{bmatrix}
\\
\textbf{J}_{t_d}  &= 
-{{\textbf{R}^I_C}{^\mathsf{T}}}
{\check{\textbf{R}}^G_{I_j}}{^\mathsf{T}}
        \check{\textbf{R}}^G_{I_i}
\begin{bmatrix}
        ^{I_i}\check{\textbf{P}}_{f_k}
\end{bmatrix}_{\times}{\boldsymbol{\omega}_i}  \\
&\quad +{{\textbf{R}^I_C}{^\mathsf{T}}}
{\check{\textbf{R}}^G_{I_j}}{^\mathsf{T}}(
        {^G\textbf{v}_{I_i}} - {^G\textbf{v}_{I_j}}
) \\
&\quad +{{\textbf{R}^I_C}{^\mathsf{T}}}
\begin{bmatrix}
        ^{I_j}\check{\textbf{P}}_{f_k}
\end{bmatrix}_{\times}{\boldsymbol{\omega}_j}\\
{^{I_i}\check{\textbf{P}}_{f_k}} &=  
\textbf{R}^I_C{^{C_i}\textbf{P}_{f_k}} + {^{I}\textbf{P}_C}
\\
{^{I_j}\check{\textbf{P}}_{f_k}} &=      
\check{\textbf{R}}^G_{I_j}{^\mathsf{T}}(
{^G\check{\textbf{P}}_{I_i}}
{^{I_i}\check{\textbf{P}}_{f_k}}+{^G\check{\textbf{P}}_{I_i}}-{^G\check{\textbf{P}}_{I_j}})
\end{aligned} \label{eqn-7.2}
\end{equation}
\subsection{Optimization with Temporal Calibration}
Referring to~\cite{qin2018vinsmono}, the overall cost function is defined as Eq.~{(\ref{eqn-9})}
\begin{equation}
        \min\limits_{\textbf{X}}
        \begin{Bmatrix}
                \\
                {\underbrace{\left|\left| {\textbf{r}_p - \textbf{J}_p\textbf{X}} \right|\right|^2}_{prior\ factor}} +
                {\underbrace{{{\sum\limits_{i \in\mathcal{B} }} \left|\left| \textbf{r}_i
                \right|\right|^2_{\textbf{P}^i_{i+1}}}}_{inertial\ factor}} +
                {\underbrace{{{\sum\limits_{{(i,j)}\in\mathcal{C} }} \left|\left| \textbf{r}_{i,j}
                \right|\right|^2_{\textbf{P}^j_{i}}}}_{visual\ factor}}
        \end{Bmatrix} \label{eqn-9}
\end{equation}
where, $\textbf{r}_p$ and $\textbf{J}_p$ are the prior residual and prior Jacobian of marginalization information, respectively. 
$\textbf{P}^i_{i+1}$ is the covariance pre-integration. 
$\textbf{P}^j_i$ is the covariance of observation noise of the $j^{th}$ feature on the $i^{th}$ window. $\mathcal{B}$ is the set of IMU pre-integrations, and $\mathcal{C}$ is the set of feature observations. The details of this formulation can be found in~\cite{qin2018vinsmono}. Finally, the cost function and its corresponding Jacobian can be substituted into general optimization tools, like g2o~\cite{kummerle2011g}, ceres~\cite{Agarwal_Ceres_Solver_2022} or other solvers to minimize the cost function and calculate the corresponding parameters.

\section{EXPERIMENTS}
Review of well-known optimization-based VINS frameworks reveals that VINS-Mono~\cite{qin2018vinsmono} is the only one with an online temporal calibration mechanism.
Therefore, our method is primarily integrated into VINS-Mono to replace its original temporal calibration module. The modified VINS-Mono is referred to as 'Ours', and the original VINS-Mono with temporal calibration enabled is referred to as 'VM'. Comparison experiments are then conducted on real-world and simulated datasets using these two implementations.
\begin{center}
        \begin{table}[t]
        \caption{Capability comparison of online temporal calibration configured with different challenging initial time offset. The unit is millisecond.}
        \label{tab:offset_comparison}
          \centering
          \begin{threeparttable}
              \scriptsize
             \resizebox{\linewidth}{!}{
             \begin{tabular}{l|ll|ll|ll}
                \toprule
                            &  Ours &  VM & Ours & VM  &  Ours &  VM  \\
                \midrule
                Offset      & 30.00 & 30.00 & 60.00 & 60.00 & 90.00 & 90.00 \\
                \midrule
                MH01        & 29.87 & 29.87 & 59.85 & 59.85 & 89.83 & 89.80 \\
                MH02        & 29.97 & 30.03 & 59.93 & 60.04 & 89.92 & 90.02 \\
                MH03        & 29.91 & 29.94 & 59.89 & 59.95 & 89.85 & 89.95 \\
                MH04        & 29.88 & 29.91 & 59.87 & 59.90 & 89.72 & 89.81 \\
                MH05        & 29.98 & 30.04 & 59.91 & 60.07 & 89.92 & 90.00  \\
                V11         & 29.63 & 30.19 & 58.55 & 60.13 & 89.27 & 90.07  \\
                V12         & 29.88 & 30.10 & 59.84 & 60.03 & 89.88 & 89.98 \\
                V13         & 29.78 & 29.89 & 59.69 & 59.89 & 89.77 & 89.92  \\
                V21         & 29.84 & 30.02 & 59.57 & 60.00 & 89.73 & 90.02  \\
                V22         & 29.97 & 29.95 & 59.97 & 59.95 & 89.98 & 89.94 \\
                \midrule
                Average     & 29.87 & \textbf{29.99} & 59.71 & \textbf{59.98} & 89.79 & \textbf{89.95} \\
                \bottomrule     
             \end{tabular}
             }
          \end{threeparttable}
        \end{table}
\end{center}

\begin{center}
        \begin{table}[t]
        \caption{RMSE evaluation of compared algorithms configured with different challenging initial time offset. Time offsets are reported in milliseconds, whereas other values are reported in centimeters.}
        \label{tab:rmse_euroc}
          \centering
          \begin{threeparttable}
              \scriptsize
             \resizebox{\linewidth}{!}{
                \begin{tabular}{l|rr|rr|rr}
                \toprule
                            & Ours & VM & Ours & VM & Ours & VM \\
                \midrule
                Offset      & 30 & 30& 60 & 60 & 90 & 90 \\
                \midrule
                MH01        & 18.2 & 15.7  & 18.8 & 15.7   & 18.7 & 16.4 \\
                MH02        & 17.9 & 17.8  & 17.5 & 17.7   & 17.9 & 17.9 \\
                MH03        & 19.9 & 19.5  & 20.1 & 19.5   & 25.0 & 19.5 \\
                MH04        & 32.3 & 31.0  & 32.3 & 44.0   & 35.3 & 30.7 \\
                MH05        & 30.3 & 28.2  & 31.3 & 28.2   & 31.1 & 32.3  \\
                V11         &  8.1 &  8.1  & 10.3 &  8.1   & 8.8  &  8.2  \\
                V12         & 15.9 & 21.4  & 15.7 & 16.4   & 13.5 & 13.7 \\
                V13         & 19.9 & 19.2  & 19.3 & 18.6   & 18.8 & 18.8  \\
                V21         &  8.3 &  9.4  &  8.2 &  8.6   & 8.4  &  9.3  \\
                V22         & 16.9 & 15.9  & 23.0 & 17.0   & 30.6 & 20.7 \\
                \midrule
                Average     & 18.77& \textbf{18.62} & 19.65& \textbf{19.38}  & 20.81& \textbf{18.75}\\
                \bottomrule     
             \end{tabular}
             }
          \end{threeparttable}
        \end{table}
\end{center}
\begin{figure}[t]
        \centering
        \includegraphics[width=\linewidth]{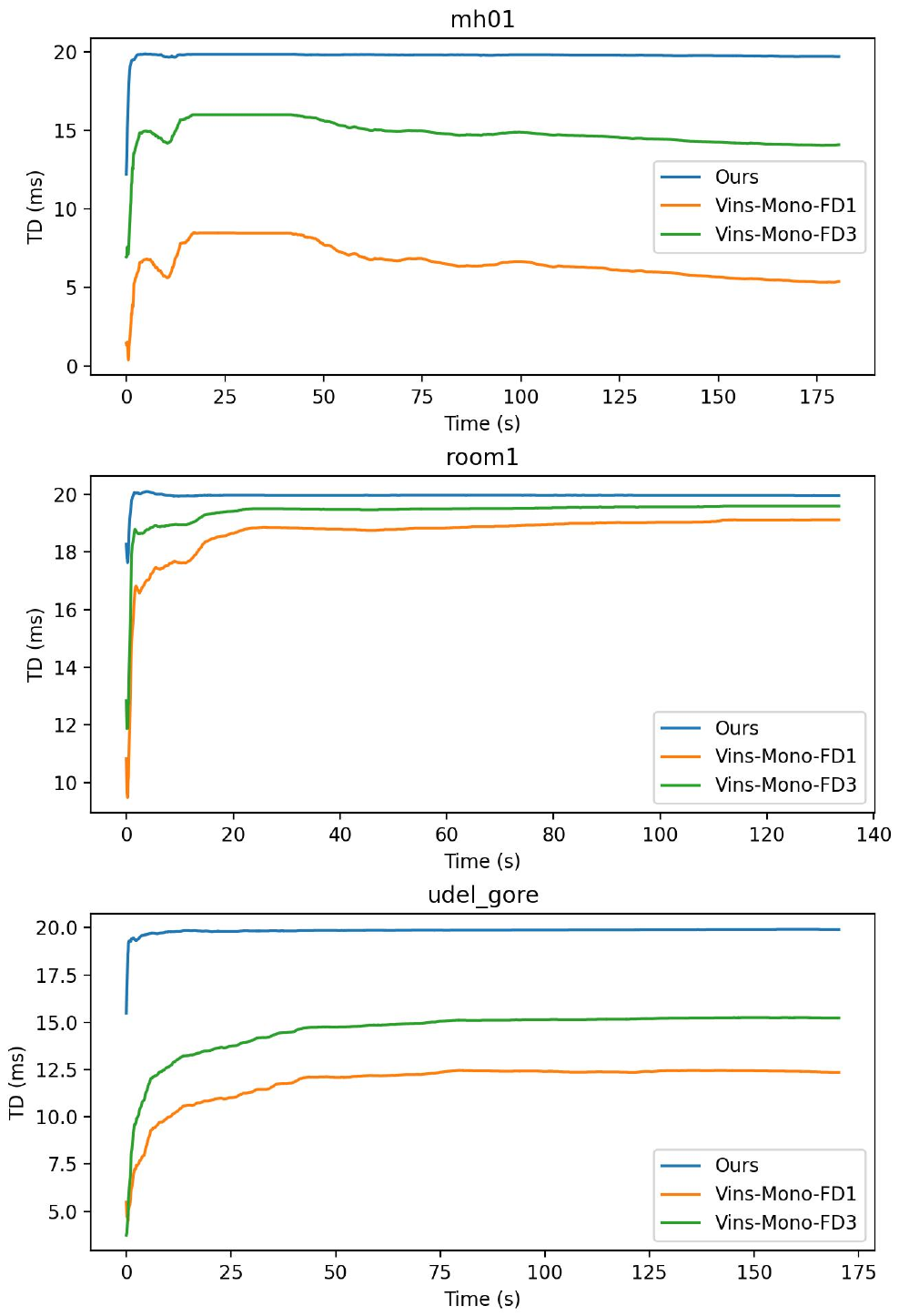}
        \caption{Time offset estimation in simulation with 20ms temporal offset. Estimated offset converges to more accurate value quickly within a few seconds.} \label{fig:td_comparison_figure}
\end{figure}
\subsection{Experiments on Real Data}
The proposed method is evaluated on EuRoC datasets~\cite{burri2016eurocdata}, which are collected onboard a micro aerial vehicle mounted the sensor suite consisting of synchronized IMU and stereo camera. The ground truth data is provided by VICON and Leica MS50. 
In this study, the image of left camera is adopted. 

Experiments on the EuRoC dataset initialize both methods with the same artificial time offset (e.g., 30, 60, or 90 milliseconds), and other configurations are kept as the default settings of the original VINS-Mono.
Each method then jointly estimates the unknown time offset and 6-DoF trajectory online. 
The final estimated time offsets are shown in Table~\ref{tab:offset_comparison} and the Root Mean Square Error (RMSE) comparisons of estimated trajectories are shown in Table~\ref{tab:rmse_euroc}.

As shown in Tables~\ref{tab:offset_comparison} and~\ref{tab:rmse_euroc}, both our method and the original VINS-Mono can successfully estimate the unknown time offset and achieve accurate 6-DoF trajectory estimation for systems with challenging time offsets. However, VINS-Mono with disabled temporal calibration cannot perform well under any  of the configurations listed in Table~\ref{tab:offset_comparison}. 
In contrast, the temporal calibration capability of the original VINS-Mono is superior to that of the proposed method. Specifically, from the formulations of visual residuals in our method, it is evident that system velocity and angular velocity are employed to model pose compensation. This approach, however, limits the immediate precision of temporal calibration because the system velocity can only gradually converge to the accurate value. Furthermore, the inaccuracy of the initial system velocity is retained in the system's marginalization information throughout the process.
\begin{center}
        \begin{table}[t]
        \caption{Capability comparison of online temporal calibration with noisy visual observations under different initial time offsets in simulation. The unit is millisecond.}
        \label{tab:simulation_offset_comparison}
          \centering
          \begin{threeparttable}
              \scriptsize
             \resizebox{\linewidth}{!}{
             \begin{tabular}{l|l|rrr|r}
                \toprule
                & Offset & MH01 & room1 & Udel$\_$gore & Average  \\
                \midrule
                Ours       & 20.0 & 19.71 & 19.95 & 19.91 & \textbf{19.86} \\
                VM-Fd1     & 20.0 & 5.36 & 19.11 & 12.36 & 12.28 \\
                VM-Fd3     & 20.0 & 14.08 & 19.59 & 15.24 & 16.30 \\
                \midrule
                Ours       & 40.0 & 39.66 & 39.91 & 39.64 & \textbf{39.74} \\
                VM-Fd1     & 40.0 & 21.96 & 39.04 & 30.53 & 30.51 \\
                VM-Fd3     & 40.0 & 33.80 & 39.57 & 33.22 & 35.53 \\
                \midrule
                Ours       & 60.0 & 59.62 & 59.86 & 59.71 & \textbf{59.73} \\
                VM-Fd1     & 60.0 & 37.99 & 58.91 & 47.57 & 48.16 \\
                VM-Fd3     & 60.0 & 53.41 & 59.36 & 48.87 & 53.88 \\
                \bottomrule     
             \end{tabular}
             }
          \end{threeparttable}
        \end{table}
\end{center}
\begin{center}
        \begin{table}[t]
        \caption{Evaluation of RMSE for various algorithms with different initial time offsets and noisy visual observations in simulation. Time offsets are reported in milliseconds, whereas other values are reported in centimeters.}
        \label{tab:simulation_rmse}
          \centering
          \begin{threeparttable}
              \scriptsize
             \resizebox{\linewidth}{!}{
             \begin{tabular}{l|l|rrr|r}
                \toprule
                & Offset & MH01 & room1 & Udel$\_$gore & Average\\
                \midrule
                Ours       & 20.0 & 12.0 & 14.2 & 20.6 & \textbf{15.60}  \\
                VM-Fd1     & 20.0 & 19.5 & 19.9 & 60.2 & 33.20  \\
                VM-Fd3     & 20.0 & 14.0 & 15.5 & 40.4 & 23.30  \\
                \midrule
                Ours       & 40.0 & 12.3 & 13.8 & 22.5 & \textbf{16.20}  \\
                VM-Fd1     & 40.0 & 21.0 & 21.3 & 70.3 & 37.53  \\
                VM-Fd3     & 40.0 & 14.3 & 15.7 & 53.5 & 27.83  \\
                \midrule
                Ours       & 60.0 & 12.2 & 14.1 & 24.5 & \textbf{16.93} \\
                VM-Fd1     & 60.0 & 31.1 & 24.5 & 101.3& 52.30 \\
                VM-Fd3     & 60.0 & 15.1 & 18.5 & 91.3 & 41.63 \\
                \bottomrule     
             \end{tabular}
             }
          \end{threeparttable}
        \end{table}
\end{center}
\subsection{Simulation Experiments}
To evaluate the robustness of the proposed methods, we conducted a series of simulation experiments. The camera operates at 30 Hz, while the IMU operates at 1000 Hz. One-pixel noise and IMU noise, as defined in Eq.~(\ref{eqn-imu_noise}), were added to the simulations.
\begin{equation}
        \begin{bmatrix}
                a_n  \\ 
                g_n  \\ 
                a_w  \\ 
                g_w
        \end{bmatrix} = 
        \begin{bmatrix}
                2.0e^{-3} \\
                1.6968e^{-4} \\
                3.0e^{-3} \\
                1.9393e^{-5}      
        \end{bmatrix}\label{eqn-imu_noise}
\end{equation}
where $a_n$ and $g_n$ are the noise density of accelerometer and gyroscope, respectively. And $a_w$ and $g_w$ are their respective random walk. 
The simulation data is generated in three scenarios (e.g., MH01~\cite{burri2016eurocdata}, room1~\cite{tum_dataset2018tum} and udel$\_$gore~\cite{geneva2020openvins}). 
To generate the interframe feature velocity for the original VINS-Mono, the forward difference method is employed with two configurations: a single step, referred to as 'Fd1', and three steps, referred to as 'Fd3'. The key difference between these two configurations is that 'Fd1' will introduce more noise into the interframe feature velocity compared to 'Fd3'. This increased noise in 'Fd1' would lead to reduced accuracy in time offset and pose estimation.

\begin{figure}[t]
        \centering
        \includegraphics[width=\linewidth]{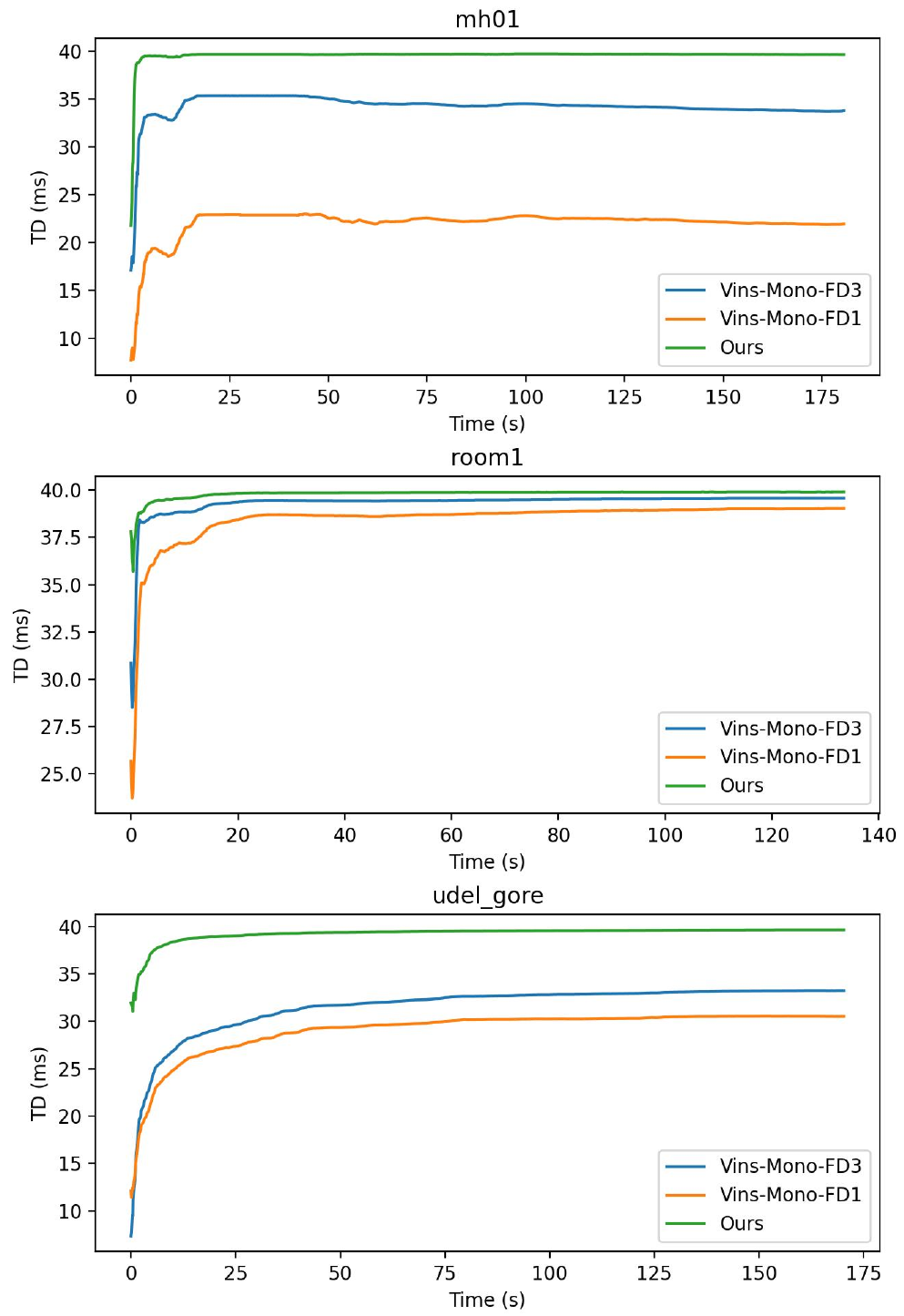}
        \caption{Time offset estimation in simulation with 40ms temporal offset. Estimated offset converges to more accurate value quickly within a few seconds.} \label{fig:td40_comparison_figure}
\end{figure}

As shown in Tables~\ref{tab:simulation_offset_comparison} and~\ref{tab:simulation_rmse}, the traditional time offset estimation capability in VINS-Mono is significantly compromised by feature tracking noise. This compromise manifests as erroneous time offsets and 6-DoF estimation inaccuracies. As illustrated in~\cref{fig:td_comparison_figure} and~\cref{fig:td40_comparison_figure}, with noisy observations, our method can quickly converge $t_d$ to an accurate value. In contrast, original VINS-Mono cannot converge to an equally accurate value. However, experiments indicate that, with the same noisy measurements, more accurate interframe feature velocity results in more accurate convergence by original VINS-Mono. Specifically, in conventional VINS-Mono, feature tracking noise leads to improper modeling of visual measurements with respect to the time offset.
Our novel method addresses these shortcomings by integrating estimated system velocity and IMU angular velocity into the visual measurement model. This integration makes our method insensitive to feature tracking noise and enables construction of a more precise and robust time offset estimation model. Consequently, our method enhances both robustness and accuracy of time offset estimation and 6-DoF estimation, significantly improving system performance.

\section{CONCLUSIONS}
In this study, we present a universal and robust online approach for temporal calibration, which does not require additional inputs like interframe feature velocity, distinguishing it from previous methods. The approach can be seamlessly integrated into any IMU-aided optimization-based sensor fusion framework including popular VINS algorithms~\cite{okvis_2015keyframe, basalt2019,dmvio2022, fan2023schurvins}, mainstream VI-SLAM~\cite{orbslam3}, and Lidar Inertial Odometry (LIO)~\cite{ye2019tightly_liom, shan2020liosam}. This integration effectively addresses misalignment of time axes among multiple sensors.
Extensive accuracy experiments demonstrate the method achieves highly accurate online calibration of time offsets and concurrent ego-motion estimation. Moreover, since the method operates within the same dimensional space as the original VINS-Mono with time offset enabled, it exhibits similar efficiency levels.
These results underscore the potential of the proposed approach to significantly enhance the reliability and accuracy of multi-sensor fusion systems across numerous applications.

\addtolength{\textheight}{-21cm}   







\bibliographystyle{IEEEtran}
\bibliography{offset_calibration}
\end{document}